\documentclass[10pt,twocolumn,letterpaper]{article}

\usepackage[pagenumbers]{cvpr} %
\usepackage[accsupp]{axessibility}  %

\usepackage{graphicx}
\usepackage{amsmath}
\usepackage{amssymb}
\usepackage{booktabs}
\usepackage{float}
\usepackage{multirow}
\usepackage{multicol}
\usepackage{dblfloatfix}
\usepackage{appendix}

\newcommand{\z}{{\rm\bf z}}                   %
\newcommand{\w}{{\rm\bf w}}                   %
\newcommand{\Loss}{\mathcal{L}}               %
\newcommand\blfootnote[1]{%
  \begingroup
  \renewcommand\thefootnote{}\footnote{#1}%
  \addtocounter{footnote}{-1}%
  \endgroup
}
\usepackage[pagebackref,breaklinks,colorlinks]{hyperref}

\usepackage[capitalize]{cleveref}
\crefname{section}{Sec.}{Secs.}
\Crefname{section}{Section}{Sections}
\Crefname{table}{Table}{Tables}
\crefname{table}{Tab.}{Tabs.}

\begin{document}

 \title{TAPS3D: Text-Guided 3D Textured Shape Generation from Pseudo Supervision}

\author{Jiacheng Wei\textsuperscript{1*} \qquad Hao Wang\textsuperscript{1*} \qquad Jiashi Feng\textsuperscript{2} \qquad Guosheng Lin\textsuperscript{1$\dagger$} \qquad Kim-Hui Yap\textsuperscript{1} \\
\textsuperscript{1}Nanyang Technological University, Singapore \quad \textsuperscript{2}ByteDance \\
\small \{\tt jiacheng002@e., hao005@e., gslin@, ekhyap@\}ntu.edu.sg, \tt jshfeng@bytedance.com}

\twocolumn[{%
\maketitle
\begin{center}
\vspace{-0.1in}
    \centering
    \captionsetup{type=figure}
    \includegraphics[width=0.99\textwidth]{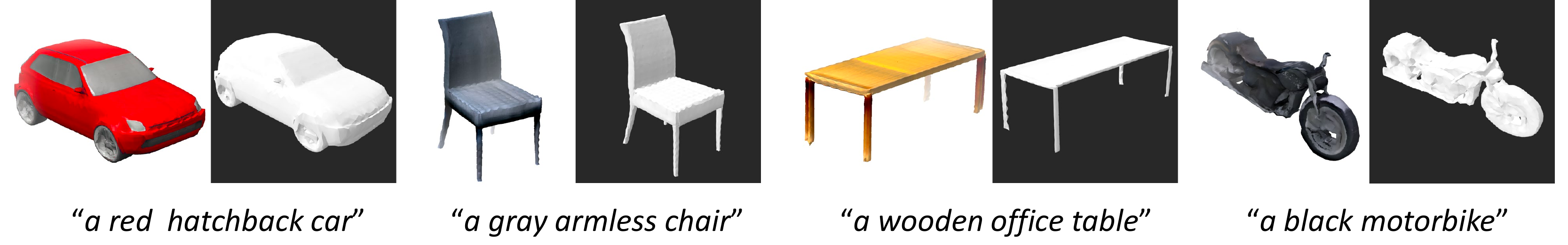}
    \captionof{figure}{Our proposed model is capable of generating detailed textured shapes based on given text prompts. We demonstrate the effectiveness of our approach by showcasing generated results for four distinct object classes: \textit{Car}, \textit{Chair}, \textit{Table}, and \textit{Motorbike}. Both the textured meshes and geometries are presented, with visualizations rendered using ChimeraX\cite{goddard2018ucsf}.}
    
    \label{fig:teaser}
\end{center}%
}]

\begin{abstract}
    \blfootnote{$^*$Equal contribution. Work done during an internship at Bytedance.}
    \blfootnote{$^\dagger$Corresponding author.}
   In this paper, we investigate an open research task of generating controllable 3D textured shapes from the given textual descriptions. Previous works either require ground truth caption labeling or extensive optimization time. To resolve these issues, we present a novel framework, TAPS3D, to train a text-guided 3D shape generator with pseudo captions. Specifically, based on rendered 2D images, we retrieve relevant words from the CLIP vocabulary and construct pseudo captions using templates. Our constructed captions provide high-level semantic supervision for generated 3D shapes. Further, in order to produce fine-grained textures and increase geometry diversity, we propose to adopt low-level image regularization to enable fake-rendered images to align with the real ones.   
   During the inference phase, our proposed model can generate 3D textured shapes from the given text without any additional optimization. We conduct extensive experiments to analyze each of our proposed components and show the efficacy of our framework in generating high-fidelity 3D textured and text-relevant shapes. Code is available at \url{https://github.com/plusmultiply/TAPS3D}
\end{abstract}

\section{Introduction}
\label{sec:intro}
3D objects are essential in various applications\cite{liu2022weakly, liu2022rm3d,liu2022fg, liu2023fac}, such as video games, film special effects, and virtual reality. However, realistic and detailed 3D object models are usually hand-crafted by well-trained artists and engineers slowly and tediously. 
To expedite this process, many research works \cite{chan2022efficient,gao2022get3d,poole2022dreamfusion,jain2022zero, song20213d, song2023unsupervised} use deep generative models to achieve automatic 3D object generation. However, these models are primarily unconditioned, which can hardly generate objects as humans will.

In order to control the generated 3D objects from text, prior text-to-3D generation works \cite{sanghi2022clip, sanghi2022textcraft} leverage the pretrained vision-language alignment model CLIP \cite{radford2021learning}, such that they can only use 3D shape data to achieve zero-shot learning. For example, Dream Fields \cite{jain2022zero} combines the advantages of CLIP and NeRF \cite{mildenhall2021nerf}, which can produce both 3D representations and renderings. However, Dream Fields costs about 70 minutes on 8 TPU cores to produce a single result. This means the optimization time during the inference phase is too slow to use in practice. 
Later on, GET3D \cite{gao2022get3d} is proposed with faster inference time, which incorporates StyleGAN \cite{karras2019style} and Deep Marching Tetrahedral (DMTet) \cite{shen2021deep} as the texture and geometry generators respectively. Since GET3D adopts a pretrained model to do text-guided synthesis, they can finish optimization in less time than Dream Fields. But the requirement of test-time optimization still limits its application scenarios. CLIP-NeRF \cite{wang2022clip} utilizes conditional radiance fields \cite{schwarz2020graf} to avoid test-time optimization, but it requires ground truth text data for the training purpose. Therefore, CLIP-NeRF is only applicable to a few object classes that have labeled text data for training, and its generation quality is restricted by the NeRF capacity. 

To address the aforementioned limitations,
we propose to generate pseudo captions {for 3D shape data} based on their rendered 2D images and construct a large amount of $\langle$3D shape, pseudo captions$\rangle$  as training data, such that the text-guided 3D generation model can be  trained over them. To this end, we propose a novel framework for \textbf{T}ext-guided 3D textured sh\textbf{A}pe generation from \textbf{P}seudo \textbf{S}upervision (TAPS3D), in which we can generate high-quality 3D shapes without requiring annotated text training data or test-time optimization.

Specifically, our proposed framework is composed of two modules, where the first   generates pseudo captions for 3D shapes  and feeds them into a 3D generator to conduct text-guided training within  the second module. In the pseudo caption generation module, we follow the language-free text-to-image learning scheme \cite{zhou2022lafite2,liang2022mind}. We first adopt the CLIP model to retrieve relevant words from given rendered images. Then we construct multiple candidate sentences based on the retrieved words and pick sentences having the highest CLIP similarity scores with the given images. The selected sentences are used as our pseudo captions for each 3D shape sample. 

Following the notable progress of text-to-image generation models \cite{nichol2021glide, ramesh2021zero, ramesh2022hierarchical, saharia2022photorealistic, yu2022scaling}, we use text-conditioned GAN architecture in the text-guided 3D generator training part. We adopt the pretrained GET3D \cite{gao2022get3d} model as our backbone network since it has been demonstrated to generate high-fidelity 3D textured shapes across various object classes. We input the pseudo captions as the generator conditions and supervise the training process with high-level CLIP supervision in an attempt to control the generated 3D shapes. Moreover, we introduce a low-level image regularization loss to produce fine-grained textures and increase geometry diversity. We empirically train the mapping networks only of a pretrained GET3D model so that the training is stable and fast, and also, the generation quality of the pretrained model can be preserved. 

Our proposed model TAPS3D can produce high-quality 3D textured shapes with strong text control as shown in Fig.~\ref{fig:teaser}, without any per prompt test-time optimization. Our contribution can be summarized as:
\begin{itemize}
\vspace{-5pt}

   \item We introduce a new 3D textured shape generative framework,  which can generate  high-quality and fidelity 3D shapes without requiring   paired text and 3D shape training data.
    \vspace{-5pt}
    \item We propose a simple pseudo caption generation method that enables text-conditioned 3D generator training, such that the model can generate text-controlled 3D textured shapes without test time optimization, and   significantly  reduce the    time cost.
    \vspace{-5pt}
    \item We introduce a low-level image regularization loss on top of the high-level CLIP loss in an attempt to produce fine-grained textures and increase geometry diversity.
    \vspace{-5pt}
 
\end{itemize}

\section{Related Work}
\subsection{Text-Guided 3D Shape Generation}
Text-guided 3D shape generation aims to generate 3D shapes from the textual descriptions so that the generation process can be controlled. There are mainly two   categories of methods, i.e., fully-supervised and optimization-based methods.
The fully-supervised method \cite{chen2018text2shape, liu2022towards, fu2022shapecrafter} uses ground truth text and the paired 3D objects with explicit 3D representations as training data. 
Specifically, CLIP-Forge \cite{sanghi2022clip} uses a two-stage training scheme, which consists of shape autoencoder training, and conditional normalizing flow training. 
VQ-VAE \cite{sanghi2022textcraft} performs zero-shot training with 3D voxel data by utilizing the pretrained CLIP model \cite{radford2021learning}.

Regarding the optimization-based methods \cite{jain2022zero, poole2022dreamfusion, michel2022text2mesh, lee2022understanding}, Neural Radiance Fields (NeRF) are usually adopted as the 3D generator. To generate 3D shapes for each input text prompt, they use the CLIP model to supervise the semantic alignment between rendered images and text prompts. Since NeRF suffers from extensive generation time, 3D-aware image synthesis \cite{chan2021pi, chan2022efficient, gu2021stylenerf, or2022stylesdf, chan2022efficient, niemeyer2021giraffe} has become popular, which generates multi-view consistent images by integrating neural rendering in the Generative Adversarial Networks (GANs). Specifically, there are no explicit 3D shapes generated during the process, while the 3D shapes can be extracted from the implicit representations, such as the occupancy field or signed distance function (SDF), using the marching cube algorithm. 
These optimization-based methods provide a solution to generate 3D shapes, but their generation speed was compromised. 
Although \cite{wang2022clip} is free from test-time optimization, it requires text data for training, which limits its applicable 3D object classes. 
Our proposed method attempts to alleviate both the paired text data shortage issues and the long optimization time in the previous work. We finally produce high-quality 3D shapes to bring our method to practical applications.

\begin{figure*}
\begin{center}
\includegraphics[width=0.9\textwidth]{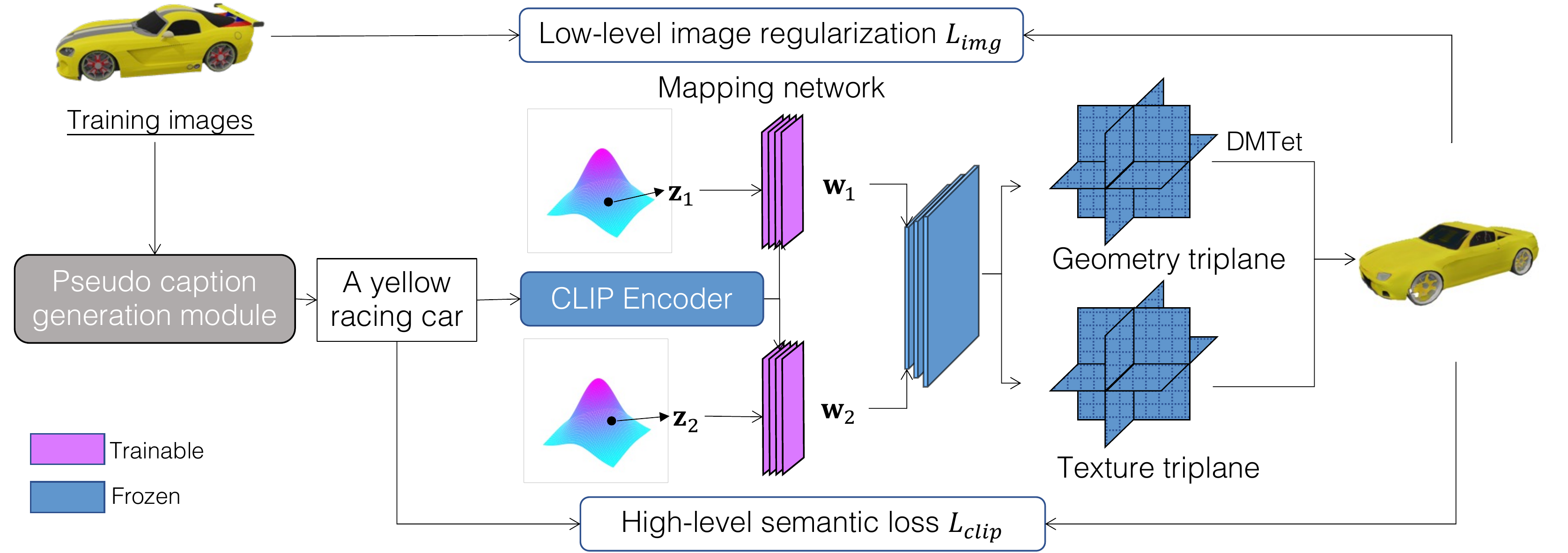}
\end{center}
\vspace{-15pt}
  \caption{Our proposed framework supports text-guided 3D shape generation training without paired text labeling. To generate captions for training, we first adopt a pseudo caption generation module to produce textual descriptions, given rendered 2D images. Then we feed the pseudo caption CLIP embeddings into the mapping networks to train the controllable text-guided 3D shape synthesis network. Our training pipeline is supervised by high-level semantic loss $\Loss_{clip}$ and low-level image regularization $\Loss_{img}$. It is notable that during the training phase, only the mapping network weights are updated, and the rest components are fixed.}
\label{fig:framework}
\vspace{-15pt}
\end{figure*}

\subsection{Text-to-Image Synthesis}
We may draw inspiration from text-to-image generation methods. Typically, many research works \cite{reed2016generative, zhang2017stackgan, qiao2019mirrorgan} adopt the conditional GAN architecture, where they directly take the text features and concatenate them with the random noise as the input. Recently, autoregressive models \cite{ramesh2021zero,ding2021cogview} and diffusion models\cite{nichol2021glide, ramesh2022hierarchical, rombach2022high, saharia2022photorealistic} made great improvement on text to image synthesis while demanding huge computational resources and massive training data.

With the introduction of StyleGAN \cite{karras2019style, karras2020analyzing, karras2021alias} mapping networks, the input random noise can be first mapped to another latent space that has disentangled semantics, then the model can generate images with better quality. Further, exploring the latent space of StyleGAN has been proved useful by several works \cite{patashnik2021styleclip, abdal2022clip2stylegan, kocasari2022stylemc} in text-driven image synthesis and manipulation tasks, where they utilize the pretrained vision-language model CLIP \cite{radford2021learning} model to manipulate pretrained unconditional StyleGAN networks. In order to relieve the need for paired text data during the training phase, Lafite \cite{zhou2021lafite} proposes to adopt the image CLIP embeddings as training input while using text CLIP embedding during the inference phase.
However, as discussed in the recent works \cite{zhou2022lafite2,liang2022mind}, there exists a modality gap between text and image embeddings in the CLIP latent space. Therefore, language-free methods that are solely trained on CLIP image embedding could introduce distribution mismatch when encountering the CLIP text embedding input during inference. In this paper, we further explore language-free training for text-guided 3D textured shape generation by proposing a pseudo caption generation module.

\section{Method}

\subsection{Preliminary}
GET3D \cite{gao2022get3d} is a newly proposed 3D generator that can generate high-fidelity textured 3D shapes. Specifically, GET3D maps noise vectors $\z \in {\mathcal N}(0, \rm\bf I)$ to a textured mesh. In order to disentangle the geometry and texture information, GET3D takes two random vectors $\z_1$ and $\z_2$ as inputs for the geometry and texture generation branches, respectively. Following the design of StyleGAN \cite{karras2019style}, and EG3D \cite{chan2022efficient}, they map $\z_1$ and $\z_2$ to the intermediate latent codes $\w_1$ and $\w_2$ with two mapping networks since the semantics would be better represented in the mapped space. The geometry and texture generators are built conditioned on $\w_1$ and $\w_2$.  

\subsection{Overview}
Since the GET3D \cite{gao2022get3d} model only supports unconditional 3D generation, if one need to do text-guided 3D shape synthesis, additional optimization is required with the original GET3D model. In order to control the semantics of the generated results and achieve text-based textured 3D generation in one shot, we propose to add text conditions to GET3D models. In \cref{fig:framework}, we present our training framework.

To this end, we address two main challenges: one is the lack of training textual descriptions, and another is color and geometry distortion. Regarding the first challenge, we propose a pseudo caption generation module, which gives large amounts of captions for training purposes. These captions are used as conditions and fed into the mapping networks to control the semantic meanings of the output textured 3D shapes. We apply the high-level CLIP \cite{radford2021learning} semantic loss on the paired input text and rendered 2D images to achieve text-3D alignment. 
To resolve the second challenge, we propose to regularize the color and geometry training with low-level image regularization. This method provides complimentary image-level guidance to our learning objective, which gives fine-grained textures during the generation phase.

Notably, we empirically adopt the pretrained GET3D model and train its mapping networks only since we observe this practice not only improves the training efficiency but also stabilizes the training process.

\begin{figure}[t]
\begin{center}
\includegraphics[width=0.48\textwidth]{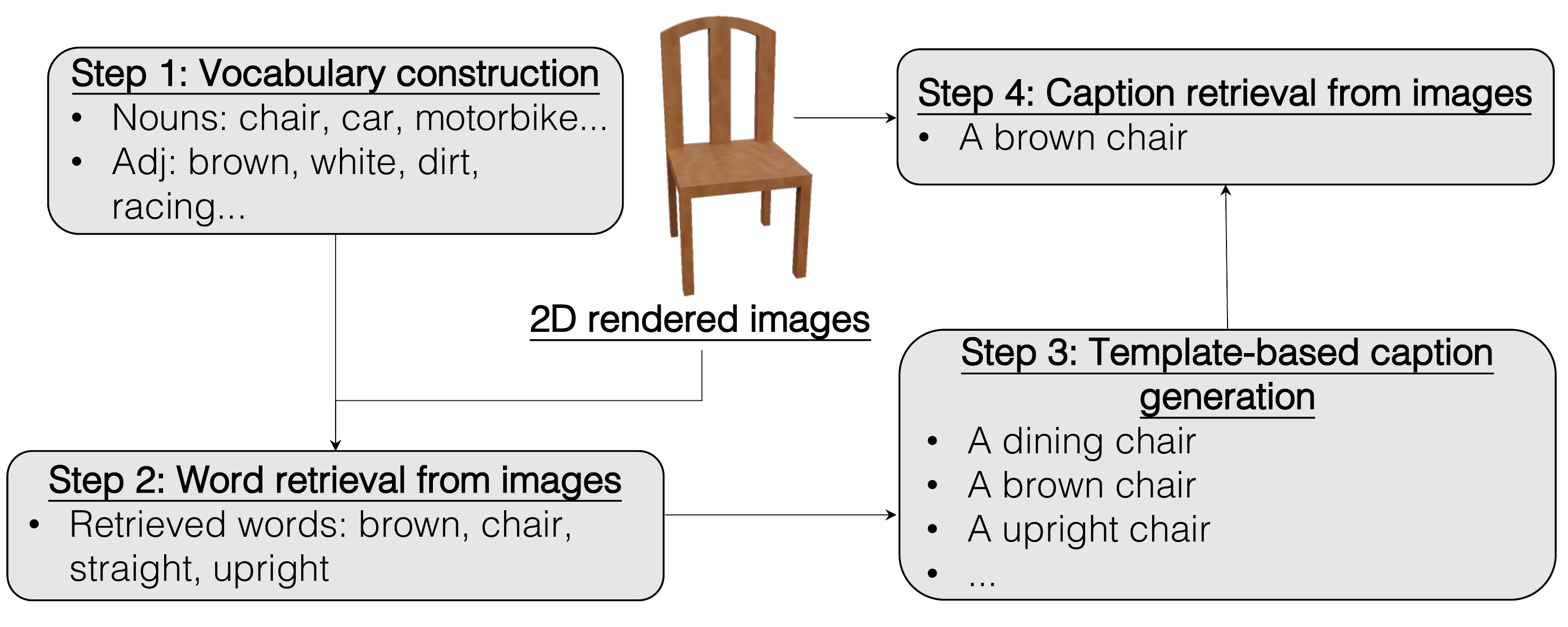}
\end{center}
\vspace{-15pt}
  \caption{Our proposed pseudo caption generation module can be split into four steps. Firstly, we construct vocabulary from ShapeNet-related nouns and adjectives in the CLIP vocabulary. Then we retrieve multiple words based on the 2D-rendered images. In the third step, we create candidate captions with retrieved words. Finally, we select one caption based on the computed text-image similarities with the CLIP model.}
\vspace{-15pt}
\label{fig:pseudo}

\end{figure}

\subsection{Pseudo Caption Generation}
Most recent 3D shape generation models are data-driven. With large-scale training data, their performance improvement is remarkable. However, when we aim to train a text-guided 3D shape generation model, we can only obtain a few textual descriptions for limited samples. To resolve this issue, we follow the method of \cite{zhou2022lafite2} and propose to generate pseudo captions for each 3D object.

Specifically, we observe the textual descriptions for a single 3D object are highly structured, e.g., \emph{``A yellow sports car''}, which typically consists of adjectives and a noun. Therefore, it is critical to obtain suitable adjectives and nouns for given 3D objects to construct pseudo captions automatically. To this end, we adopt the pretrained text-image similarity model of CLIP \cite{radford2021learning} to retrieve useful nouns and adjectives and further build complete sentences. 

The pseudo caption generation module can be split into four steps, shown in \cref{fig:pseudo}. In the first step, we construct a vocabulary to cover the related adjectives and nouns. Since we aim to generate ShapeNet \cite{chang2015shapenet} object classes, we collect all the class names as nouns in the vocabulary, which has 169 tokens. Then we extract all the adjectives from CLIP vocabulary with the Natural Language Toolkit \cite{nltk}. The size of our collected adjectives is 1463. 

In the second and third steps, we aim to retrieve relevant words from 2D-rendered images and construct candidate captions. Technically, we first extract all the vocabulary word embeddings and rendered image embeddings, where we adopt the pretrained CLIP text and image encoders, respectively. Then for each object, we compute the cosine similarities between the given image embedding and all the text embeddings. It is notable that we retrieve the nouns and adjectives separately, where we select the top-$K_1$ retrieved nouns and top-$K_2$ retrieved adjectives. With the selected words, we utilize the templates to construct candidate sentences. Since these sentences are only supposed to describe a static object, we use the template ``a \{\texttt{Adjective}\} \{\texttt{Noun}\}'', where we put 1 or 2 adjectives randomly to increase the caption diversity. Due to having $K_1$ nouns and $K_2$ adjectives, we are able to combine them and produce multiple candidate pseudo captions. 

In order to obtain the most suitable captions for given objects, we further conduct image-to-candidate caption retrieval, where we also use CLIP encoders to extract the image and sentence embeddings. To be specific, we use the top retrieved sentences as the final pseudo captions, which are used as the input condition for the 3D generator.

\begin{figure}
\begin{center}
\includegraphics[width=0.48\textwidth]{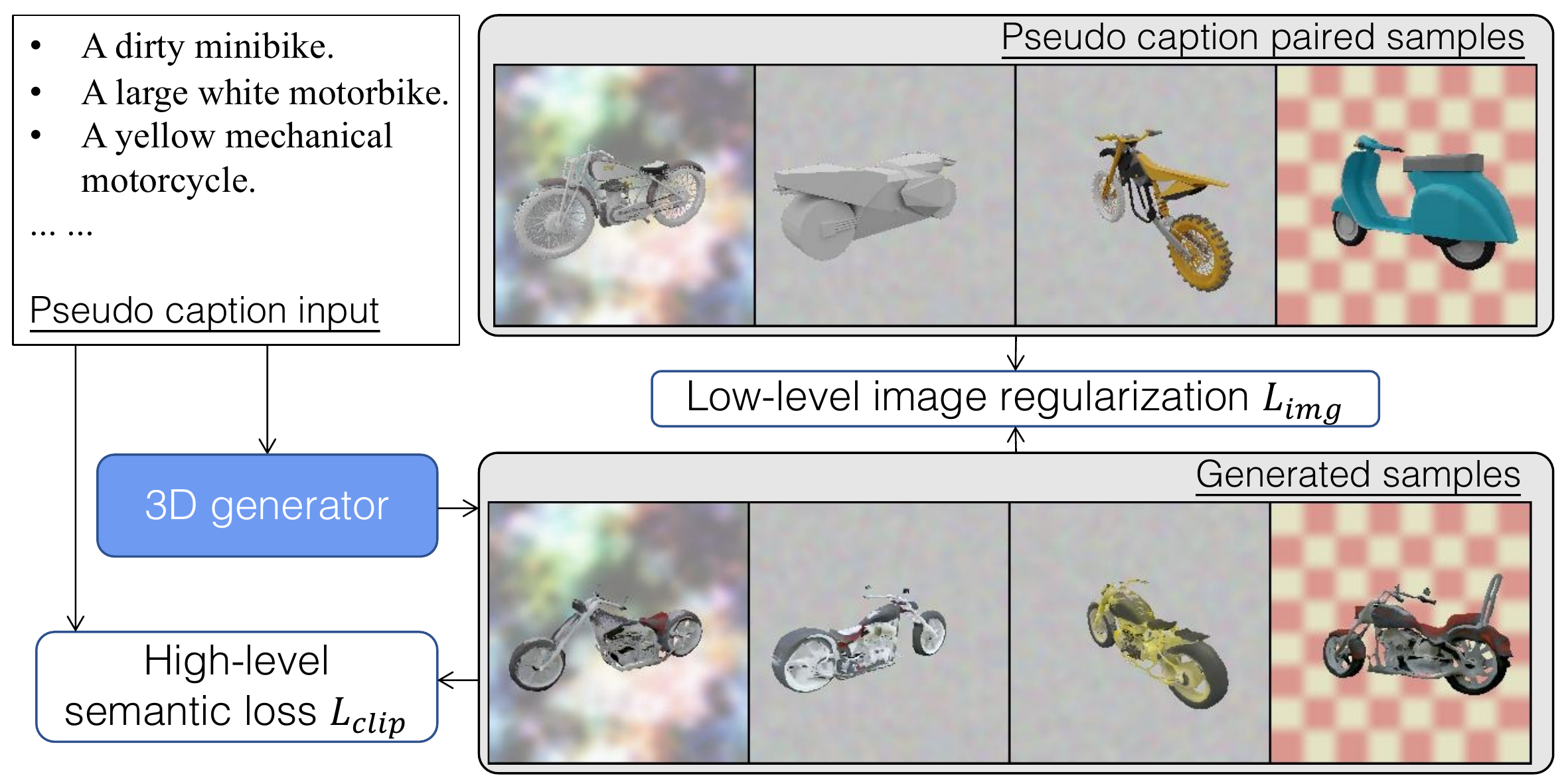}
\end{center}
\vspace{-15pt}
  \caption{In the training phase, we not only conduct the high-level text-to-image similarity learning through the CLIP model but also use low-level image regularization to obtain fine-grained textures and diverse shapes, where we render real and fake images with the same camera poses and apply background augmentation to allow the model focus on object features only.}
\vspace{-15pt}
\label{fig:img2img}
\end{figure}

\subsection{Text-3D Alignment Training}
After the pseudo caption generation, we feed these captions into a pretrained GET3D \cite{gao2022get3d} model in an attempt to control the semantic meanings of generated 3D objects. To be specific, we first adopt the CLIP text encoder $E_t$ to extract the caption $t$ features, then the caption features $E_t(t)$ are served as conditions and fed into the mapping networks of pretrained GET3D models. The input of the original mapping networks is random noise only. To inject the text information, we concatenate the caption embeddings and the random noise together as the new input. This enables us to generate diverse 3D objects that match the text's semantic meanings. 

Since we adopt the pretrained 3D generator as our model backbone, which has a good ability to synthesize high-fidelity 3D textured shapes, fixing the generator weights allows our model to focus on getting semantic alignment between input textual descriptions and generated 3D objects and also alleviates the learning difficulty. Our main training objective is to maximize the image-text similarity measured by the CLIP model, which can be denoted as:
\begin{equation}\label{eq:lossclip}
    \Loss_{clip} = 1 - \cos(E_i(I_x), E_t(t)).
\end{equation}
Here $E_i$ and $I_x$ represent the CLIP image encoder and rendered images, respectively. These 2D images are rendered from the generated 3D textured shapes with randomly sampled camera poses.

\subsection{Towards Diverse and Fine-Grained Generation}
It is observed using $\Loss_{clip}$ only enables the model to generate plausible 3D shapes that align with the input text. However, the diversity of geometry generation is limited, and the generated colors are unnatural. This is caused by two main reasons. One reason is the diversity of our pseudo captions is also limited, as there are only a few descriptive words for a single object class. Moreover, the CLIP model may wrongly retrieve irrelevant words to the given objects, which also affects the generated shape quality. Another reason is the CLIP model only provides high-level guidance for our training supervision while failing to give fine-grained texture information, thus making the generated results unrealistic.

In order to resolve this issue, we further introduce a low-level image regularization loss. Specifically, we use the same camera poses as that used in the $I_x$ rendering to produce ground truth rendered images $I_x^{gt}$. The optimization objective is denoted as 
\begin{equation}\label{eq:img2img}
    \Loss_{img} = 1 - \cos(E_i(I_x), E_i(I_x^{gt})).
\end{equation}
Since the rendering process returns an alpha channel, which results in a simple white or black background, if we directly take rendered images with the same backgrounds over training iterations, the background information will also be involved during optimizing \cref{eq:img2img}.  
To alleviate this issue and allow the model to focus on the foreground objects, we follow \cite{jain2022zero} to conduct background augmentation, as shown in \cref{fig:img2img}. Technically, we randomly select random Fourier textures \cite{mordvintsev2018differentiable}, Gaussian noise, and checkerboard patterns as backgrounds. For each paired real and fake image, we apply the same augmented backgrounds.

\subsection{Overall Training Objective}
Our overall training objective can be depicted as
\begin{equation}
    \Loss = \Loss_{clip} + \Loss_{img}.
\end{equation}
During the training phase, we empirically experiment with updating all model parameters and the mapping networks only. We observe training all GET3D backbone parameters would severely distort the generated geometry and textures, as illustrated in \cref{fig:all_vs_part}. This is because the learning objective is two-pronged: improving the generation quality and matching the input captions. While borrowing a pretrained unconditional generator ensures good generation quality, expedites the training process, and also yields stable text-guided results.
Therefore, we regard training the mapping networks only of a pretrained 3D generator as a useful and practical solution for the text-guided 3D shape generation task. 

\begin{figure}
\begin{center}
\includegraphics[width=0.45\textwidth]{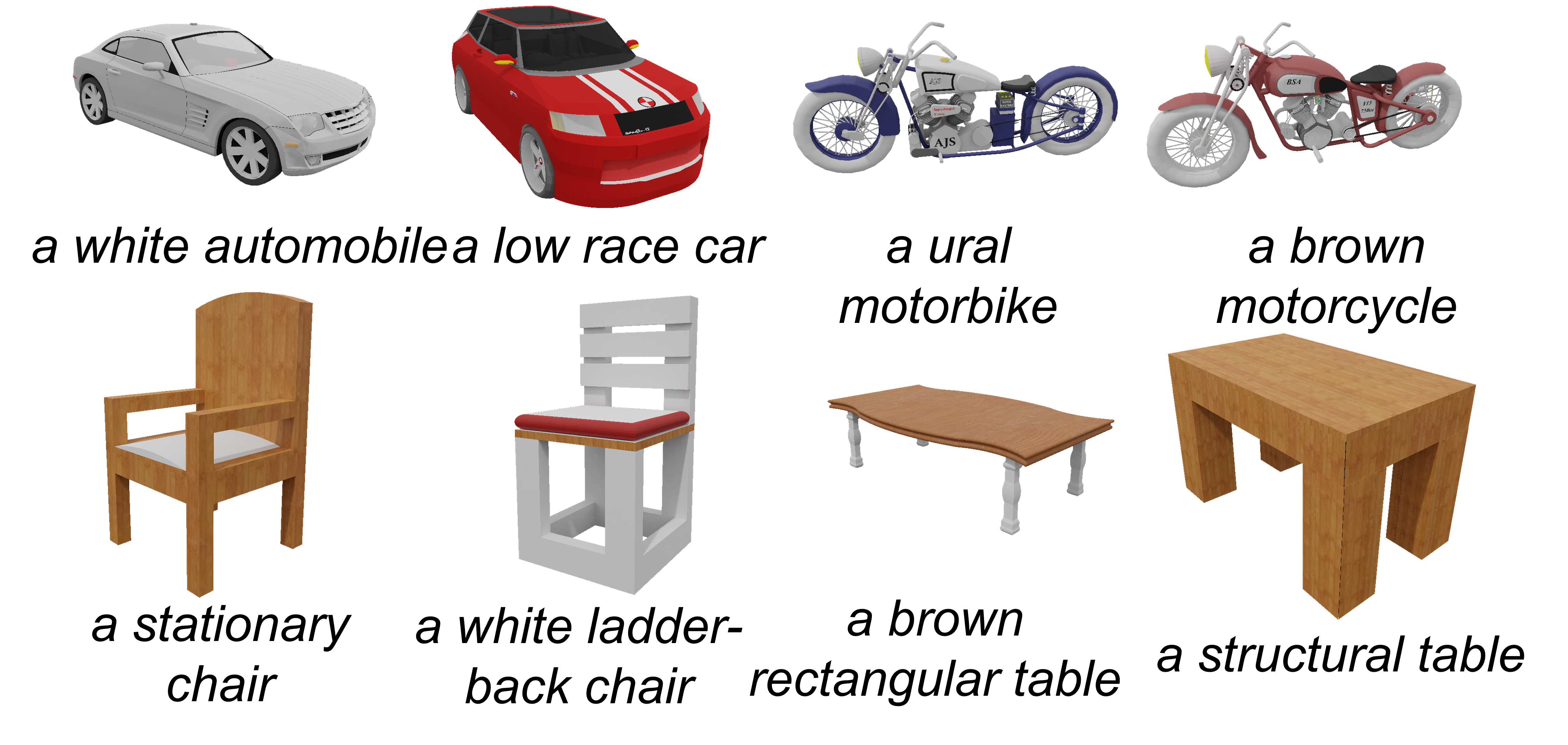}
\end{center}
\vspace{-15pt}
  \caption{The demonstration of our generated pseudo captions. }
\vspace{-15pt}
\label{fig:caption_vis}
\end{figure}

\begin{figure*}
\begin{center}
\includegraphics[width=0.9\textwidth]{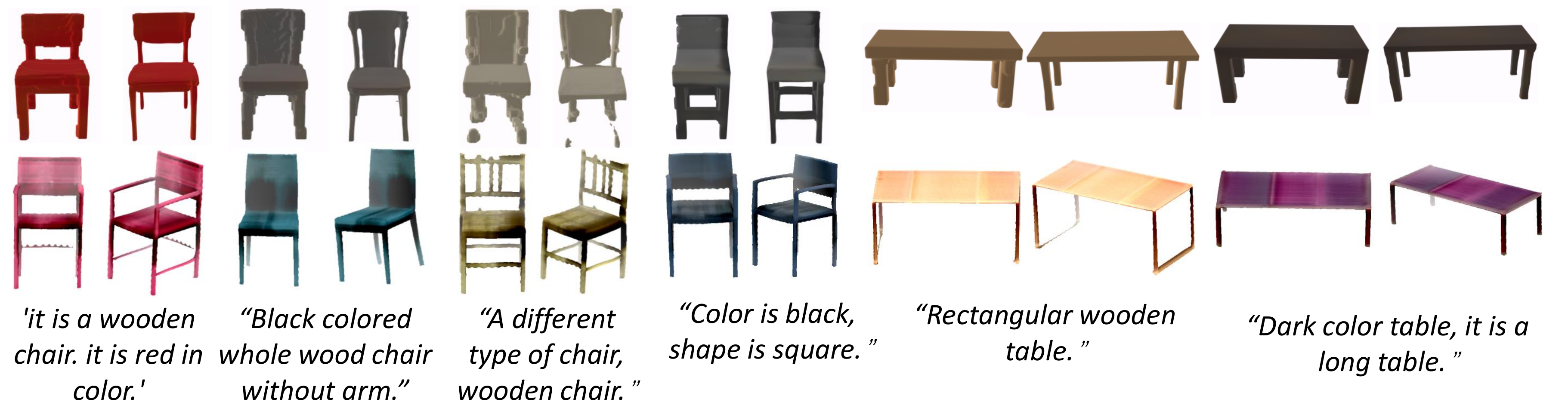}
\end{center}
\vspace{-15pt}
  \caption{We compare the text-guided generation results of our methods with that of the fully supervised method TITG3SG \cite{liu2022towards} on two object classes, \emph{Chair} and \emph{Table}.
  On the top row, we present the results from \cite{liu2022towards}, for each text prompt, the top left one is the raw voxel output in $64^3$ resolution, and the top right one is the refined mesh based on the voxel output. The bottom row shows our mesh output with two different views. The renderings are generated with ChimeraX\cite{goddard2018ucsf}.}
\vspace{-15pt}
\label{fig:comparetowards}
\end{figure*}

\section{Experiments}
\subsection{Dataset}
We train and evaluate our method on the large-scale 3D dataset ShapeNet\cite{chang2015shapenet}. Our experiments are conducted on four classes with complex geometry, \textit{Car}, \textit{Table}, \textit{Chair}, and \textit{Motorbike}, containing 7497, 8436, 6778, and 337 shapes, respectively. We follow GET3D\cite{gao2022get3d} to split the data into training, validation, and test sets in a 7:1:2 ratio. The training image data are rendered following the GET3D setting. We render 24 views randomly for \textit{Car},  \textit{Chair} and \textit{Table} and 100 views for \textit{Motorbike} since it has less available data. The pretrained models were trained on each class separately.

\subsection{Implementation details}
We use the GET3D \cite{gao2022get3d} model pretrained on ShapeNet \cite{chang2015shapenet} dataset as our backbone 3D generator. In the pseudo caption generation module, we set $K_1$ and $K_2$ as 3 and 6, respectively. We generate 20 pseudo captions for each ShapeNet object. During the training phase, each input image is randomly paired with one of the pseudo captions generated for the object which the image was rendered from. An example of the generated pseudo can be found in \cref{fig:caption_vis}. We only update the geometry and texture mapping networks. We set the batch size to 16 and ran all the experiments on 4 Nvidia Tesla V100-32G GPUs.
We empirically observe the geometry branch is more sensitive to the weight update. Hence we set learning rates for geometry and texture mapping networks as 0.004 and 0.0005, respectively. It costs around 10 hours for model training.

\subsection{Comparison with Existing Methods}
\subsubsection{Qualitative comparison}
In \cref{fig:comparetowards}, we first compare with the method TITG3SG \cite{liu2022towards}, which is an encoder-decoder-based framework. They only experiment with object classes of \textit{Chair} and \textit{Table}. Here we conduct the comparison qualitatively. Specifically, we show the generated explicit 3D shapes with the corresponding text prompts. Since TITG3SG \cite{liu2022towards} generates 3D shapes in two steps, where the coarse voxel and refined mesh are produced separately. We show them both in the top row of \cref{fig:comparetowards}.
The second row presents our generated mesh in two different views, in which we generates final 3D results in one shot. We observe our method gives better 3D textured shapes than TITG3SG, which is a fully supervised method. For example, in the third generated sample of TITG3SG, the chair legs are distorted. Moreover, the high alignment between input prompts and our generated results demonstrates our text control ability.

\subsubsection{Quantitative comparison}
To demonstrate our 3D generation quality, we compare existing works \cite{liu2022towards,wang2022clip,sanghi2022clip} from three aspects: 1) rendered 2D image quality using FID, 2) text-image relevance using R-Precision, and 3) 3D geometry quality using FPD.

In \cref{tab:existing}, we provide a quantitative comparison with CLIP-NeRF \cite{wang2022clip}, which utilizes human-written captions for training. We employ the Fr\'echet Inception Distance (FID) \cite{heusel2017gans} as our evaluation metric and present results for two object classes: \textit{Car} and \textit{Chair}. To ensure fair comparisons, we downsample our results to match the resolution of CLIP-NeRF \cite{wang2022clip}. CLIP-NeRF employs NeRF as the 3D generator, which imposes limitations on the generation resolution. In contrast, our backbone GET3D model boasts a larger capacity, supporting higher resolution. Experimental results demonstrate that our method surpasses CLIP-NeRF in text-guided generation quality.

In \cref{tab:cliprp}, we adopt the CLIP-R-Precision metric \cite{park2021benchmark} to evaluate the consistency between input text prompts and generated shapes. Given that our model is category-specific, we perform per-category CLIP-R-Precision and compare it to the fully supervised TITG3SG method \cite{liu2022towards}. According to \cite{park2021benchmark}, we compute the CLIP-R-Precision using a set of unseen random captions with words sampled from existing captions, thereby enhancing correlation with human evaluations and mitigating model bias inherent compared to the conventional R-Precision metric. It is observed our method outperforms TITG3SG \cite{liu2022towards} in each category.

In \cref{tab:fpd}, we assess the geometry generation quality using the Fr\'echet Point Distance (FPD) \cite{li2021sp,shu20193d}, which measures the feature distance of a PointNet \cite{qi2017pointnet} pretrained on ModelNet40 \cite{wu20153d}. Our method outperforms existing approaches in terms of geometry quality by a large margin.

\begin{table}
 \caption{Comparison with the existing work. We evaluate the rendered 2D images using Fr\'echet inception distance (FID). We downsample our result to the same resolution of CLIP-NeRF \cite{wang2022clip} for fair comparisons.}%
 \vspace{-10pt}
 \label{tab:existing}
 \centering
 \resizebox{0.45\textwidth}{!}{
 \begin{tabular}{lcccc}\toprule
 & \multicolumn{2}{c}{Car} & \multicolumn{2}{c}{Chair} \\
 \cmidrule(r){2-3}\cmidrule(r){4-5}
  & Resolution & FID & Resolution & FID \\\midrule
CLIP-NeRF \cite{wang2022clip}  & $256^2$ &67.8 &$128^2$ & 48.4 \\
Ours & $256^2$ & \textbf{20.1} & $128^2$ & \textbf{43.7} \\
Ours & $1024^2$ & \textbf{21.7} & $1024^2$ & \textbf{44.8} \\
\bottomrule
\end{tabular}
}
\vspace{-5pt}
\end{table}

\begin{table}[]
\centering
\caption{Per category CLIP-R-Precision score using random generated captions on categories \textit{Chair} and \textit{Table}.}
\vspace{-10pt}
\resizebox{0.75\columnwidth}{!}{%
\begin{tabular}{lcc}
\toprule
Method                                        & Chair                         & Table                         \\ \midrule
TITG3SG \cite{liu2022towards} & 9.05 $\pm$ 1.92  & 6.67 $\pm$ 3.25  \\
Ours   & \textbf{20.19 $\pm$ 1.89} & \textbf{15.21 $\pm$ 3.15} \\ \bottomrule
\end{tabular}%
}
\label{tab:cliprp}
\vspace{-5pt}
\end{table}

\begin{table}[]
\centering
\caption{Comparison of 3D generation quality in FPD score.}
\vspace{-10pt}
\label{tab:fpd}
\resizebox{0.3\textwidth}{!}{
\begin{tabular}{lcc}
\toprule
Method   & Chair           & Table         \\ \midrule
TITG3SG \cite{liu2022towards}     & 1566.76         & 1639.68  \\
CLIP-Forge \cite{sanghi2022clip}  & 825.96          & 3051.31            \\
Ours               & \textbf{342.23} & \textbf{1468.43}   \\ \bottomrule
\end{tabular}
}
\vspace{-5pt}
\end{table}

\begin{table}
\centering
\caption{The comparison of inference time per text prompt for different methods. The inference time of DreamFields\cite{jain2022zero}, DreamFusion\cite{poole2022dreamfusion}, and PureCLIPNeRF\cite{lee2022understanding} are given in their papers. For TITG3SG \cite{liu2022towards} and ours, we generate 1000 text prompts with batch size as 1, and calculate the average inference time.}
\vspace{-10pt}
\resizebox{0.48\textwidth}{!}{
\begin{tabular}{l|ccc}
\toprule
Method     & Device            & Output  &  Time \\ \midrule
DreamFields\cite{jain2022zero}  & TPU cores x8      & Rendering   & 72 min         \\
DreamFusion\cite{poole2022dreamfusion}  & TPUv4 machine     & Rendering   & 90 min         \\
PureCLIPNeRF\cite{lee2022understanding} & GTX 2080ti     & Rendering   & 20 min         \\
TITG3SG \cite{liu2022towards}       & Telsa V100-32G & Voxel       & 2.21 sec       \\
TITG3SG \cite{liu2022towards}       & Telsa V100-32G & Mesh        & 24.44 sec      \\ \midrule
Ours         & Telsa V100-32G & Rendering   & 0.05 sec     \\
Ours         & Telsa V100-32G & Mesh        & 7.09 sec       \\ \bottomrule
\end{tabular}
\label{tab:inference-time}
}
\end{table}

\begin{figure*}
\begin{center}
\includegraphics[width=0.98\textwidth]{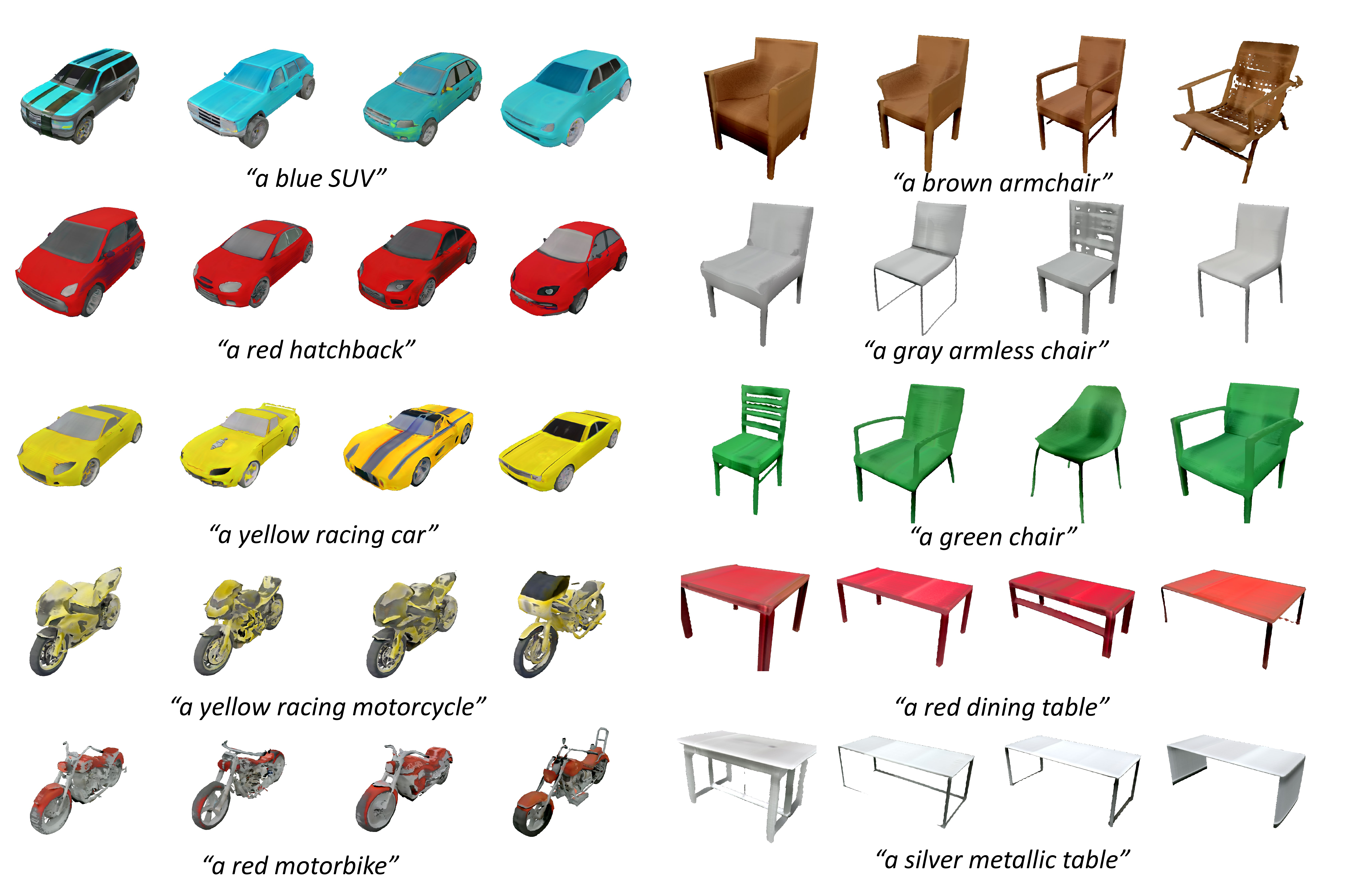}
\end{center}
\vspace{-15pt}
  \caption{We show the generation diversity and text control of our method. Each row takes the same text prompt with different sampled random noise as the model input.}
\vspace{-15pt}
\label{fig:diversity}
\end{figure*}

\subsubsection{Inference speed comparison}
In \cref{tab:inference-time}, we compare the inference speed of our method with other works. The optimization-based methods DreamFields\cite{jain2022zero}, DreamFusion\cite{poole2022dreamfusion}, and PureCLIPNeRF\cite{lee2022understanding} take tens of minutes to optimize for each input text prompt. Liu et al. \cite{liu2022towards} use an encoder-decoder architecture and thus can perform inference at 2.21 seconds. However, the raw $64^3$ voxel output is very coarse, as shown in \cref{fig:comparetowards}. The coarse output is later refined with interpolations and marching cubes to generate mesh data. This process takes ten times longer than the network inference time and costs 24.44 seconds per sample. Our method generates high-resolution renderings at $1024^2$ resolution with only 0.05 seconds and generates the complete explicit mesh representation in 7.09 seconds.

\begin{figure}
\begin{center}
\includegraphics[width=0.45\textwidth]{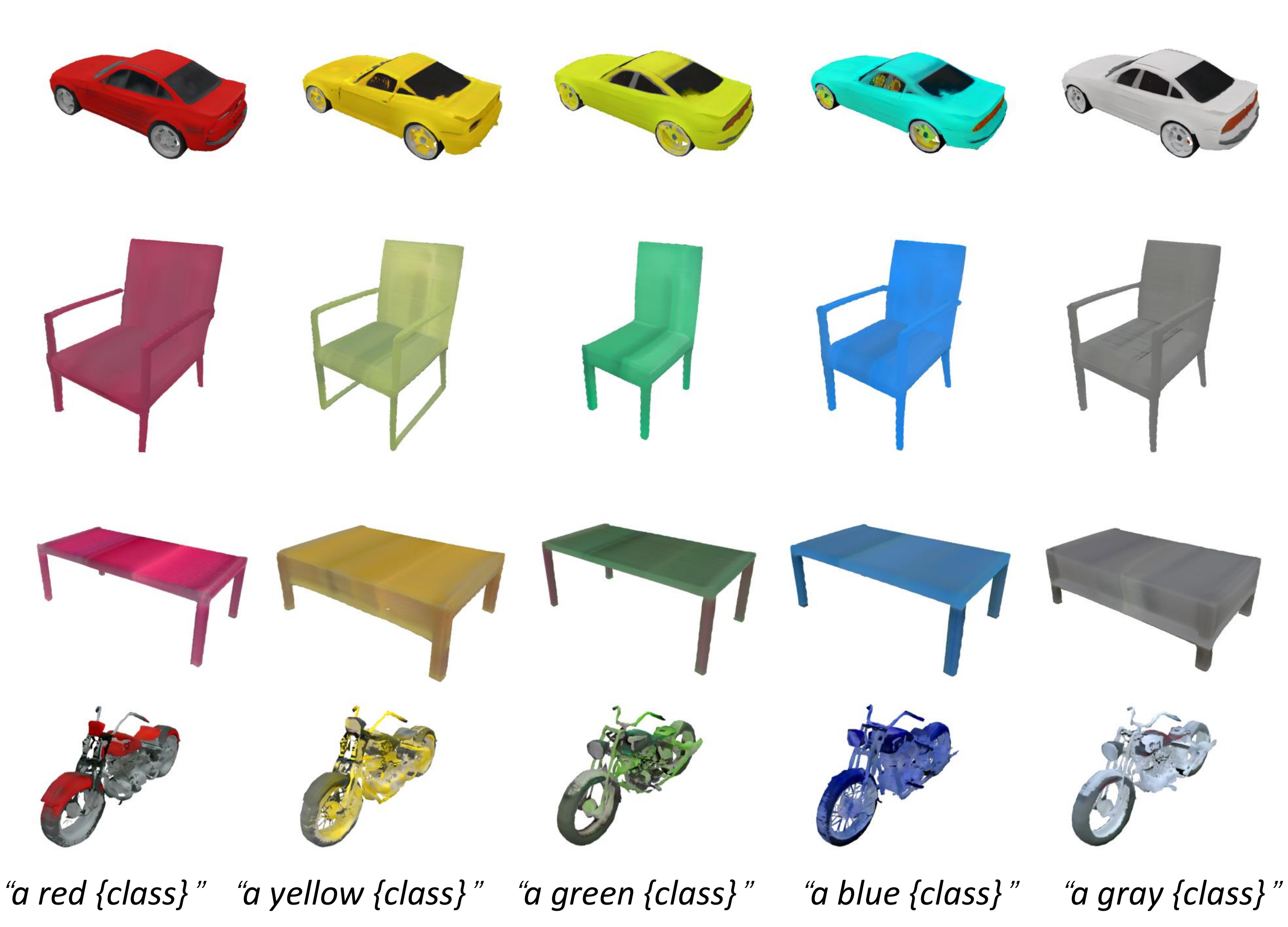}
\end{center}
\vspace{-15pt}
  \caption{Text control of textures on four different categories. Each row shares the same sampled random noise input.}
\label{fig:control}
\vspace{-15pt}
\end{figure}

\subsection{Qualitative Results}
\vspace{-5pt}
In \cref{fig:diversity}, we present the generation diversity and text control strength of our method in four object classes. In each row, we sample different random noises as well as the given text prompt as input to generate diversified and text-controlled 3D shapes, where the input text features are simultaneously fed into both the geometry and texture mapping network. Generally, we observe the semantic meanings of the rendered 2D images, and the given prompts are consistent. Moreover, there exists large diversity in the texture and geometry of our generated 3D shapes, even from the same text input. Here we present generated results of 4 classes, where \emph{Car} has a relatively simple geometry, while \emph{Table} and \emph{Chair} are more complex. But we still have good control over the textures and geometries for the challenging classes. For example, when we input \emph{``a gray armless chair''}, the generated samples exactly match the given instructions. However, we also observe failure cases of \emph{Table} generation. For instance, the generation of \emph{``a wooden office desk''} may have distorted geometry due to its complex structure. In \cref{fig:control}, we show the text control ability of our method on different classes. The results in each row are generated from the same random noise input but with different text prompts. We provide more experimental results such as interpolation and multi-view consistency in the appendix. 
To summarize, the results show that our generation results are highly consistent with the input text prompts across different object classes.

\subsection{Ablation Study}

\begin{table}
\caption{We compare the FID score of using image CLIP features (Lafite \cite{zhou2021lafite}) and the pseudo text CLIP features (Ours) as training input. }
\vspace{-10pt}
\centering
\resizebox{0.4\textwidth}{!}{
\begin{tabular}{l|cccc}
\toprule
Method & Car  & Chair & Table & Motorbike \\ \midrule
Lafite \cite{zhou2021lafite}  & 68.7 & 81.1  & 94.3  & 146.5     \\
Ours & \textbf{21.7} & \textbf{44.8}  & \textbf{43.2}  & \textbf{57.2}      \\ \bottomrule
\end{tabular}
}
\vspace{-5pt}
\label{tab:image-only}
\end{table}

\begin{table}
\caption{We compare the FID score for our method without  $\Loss_{clip}$, $\Loss_{img}$ and background augmentation (BGaug).}
\centering
\vspace{-10pt}
\resizebox{0.42\textwidth}{!}{
\begin{tabular}{l|cccc}
\toprule
Method & Car  & Chair & Table & Motorbike \\ \midrule
Ours    & \textbf{21.7} & \textbf{44.8}  & \textbf{43.2}  & \textbf{57.2}      \\ 
\quad  -w/o  $\Loss_{clip}$   &  83.7   & 148.9 & 67.1  & 134.9     \\
\quad  -w/o $\Loss_{img}$ &   59.8   & 144.2 & 48.4  & 101.8     \\
\quad  -w/o BGaug    & 55.1 &   65.1    & 58.1  & 110.0     \\
\bottomrule
\end{tabular}
}
\vspace{-20pt}
\label{tab:ablation-loss}
\end{table}

\begin{figure}
\begin{center}
\includegraphics[width=0.48\textwidth]{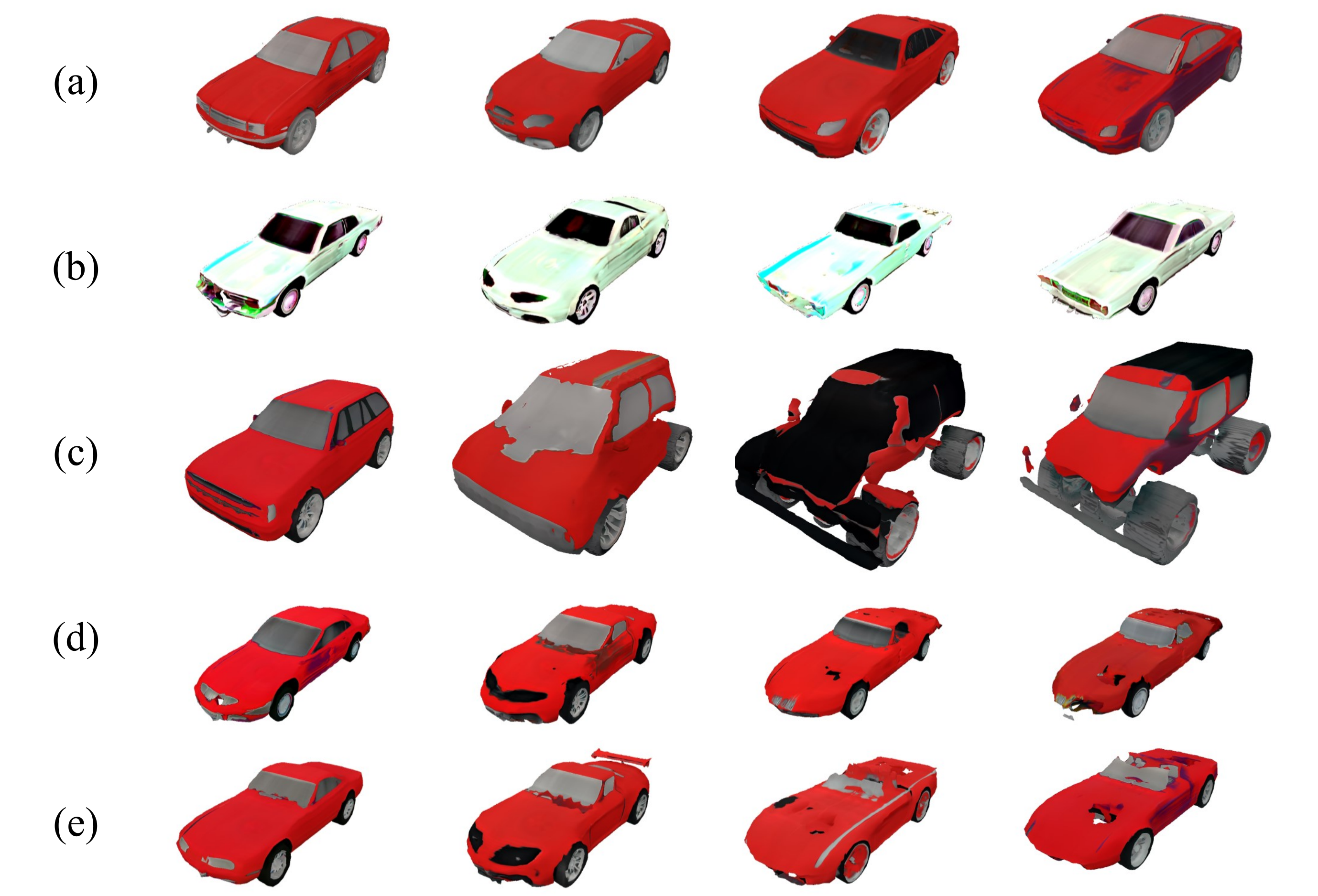}
\end{center}
\vspace{-15pt}
  \caption{Ablation study on our methods with text prompt \textit{"a red car"}. We show results of (a) our full model, (b) Lafite\cite{zhou2021lafite}, (c) without high-level semantic regularization $\Loss_{clip}$, (d) without low-level image regularization $\Loss_{img}$, (e) without background augmentation. All the visualizations are generated from models trained for 250k iterations.}
\label{fig:ablation}
\vspace{-15pt}
\end{figure}

\textbf{Pseudo caption generation module.} In \cref{tab:image-only}, we compare our method with Lafite \cite{zhou2021lafite}, which is a language-free training method for the text-to-image generation task. Lafite claims due to the alignment between CLIP text and image embeddings. The CLIP image features can be directly used as text input during training.
To implement Lafite \cite{zhou2021lafite} onto our framework, we feed the rendered real image CLIP features into our model and train the model by minimizing the cosine similarity between the generated and input image CLIP features. The quantitative results show that our proposed pseudo caption generation module significantly improves the generation quality compared to the baseline. We also give a qualitative comparison in \cref{fig:ablation} (b), we observe that the generator trained with Lafite \cite{zhou2021lafite} fails to generate correct textures. This problem might come from the modality gap between the CLIP image and text embeddings, as discussed in \cite{liang2022mind, zhou2022lafite2}.

\textbf{Effects of our loss functions.} We conduct an ablation study on our used two losses $\Loss_{clip}$ and $\Loss_{img}$. In \cref{tab:ablation-loss}, we quantitatively validate the usefulness of both two losses. Specifically, $\Loss_{clip}$ supervises the text-image alignment learning, which is more critical than $\Loss_{img}$, and removing $\Loss_{clip}$ gives a more considerable performance drop.
From the qualitative results in \cref{fig:ablation} (c), we observe that simply using $\Loss_{img}$ leads to the model generating distorted geometries, which may indicate the significance of high-level semantic supervision.
We also observe that solely using $\Loss_{clip}$ (\cref{fig:ablation} (d)) may cause the model generation to lose fine-grained texture details, as well as the geometry information.

\textbf{Effects of background augmentation.} 
Our used GET3D \cite{gao2022get3d} backbone network generates RGBA images with dark backgrounds, which may bring some noise when we conduct low-level image regularization. 
As illustrated from the quantitative results in \cref{tab:ablation-loss} and the visualizations in \cref{fig:ablation} (e), training with the augmented background can help remove the texture flaws.

\begin{figure}
\begin{center}
\includegraphics[width=0.48\textwidth]{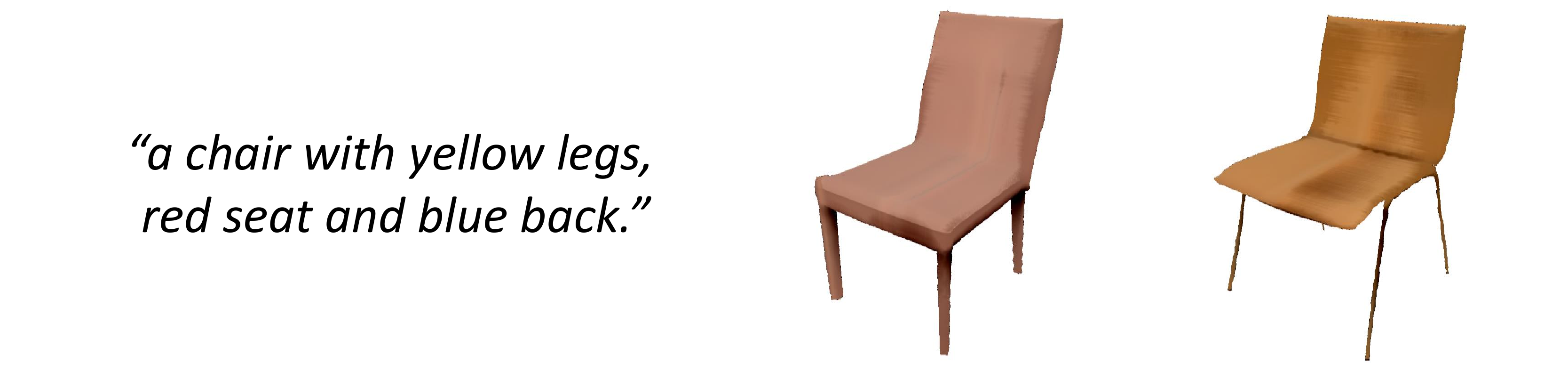}
\end{center}
\vspace{-15pt}
  \caption{Examples of failure cases of our method. Our model often fails to handle text input with composed descriptions for different parts of the object. 
}
\label{fig:failure}
\vspace{-15pt}
\end{figure}

\section{Conclusion}

We introduce TAPS3D, a text-guided 3D textured shape generation method. We first generate pseudo captions for model training. During the inference phase, TAPS3D does not require additional optimization. Our fast generation speed enables average users to generate high-quality 3D objects with acceptable processing time.

\textbf{Limitation.} The limitation of our method mainly lies in 
the capacity of our model is still bound to the training image data, and we cannot produce different fine-grained details for different object parts, as shown in \cref{fig:failure}. Although we do not require paired text data, we still need diversified training images to handle complex text input.
\vspace{-5pt}
\section*{Acknowledgment} 
\vspace{-5pt}
This research is partly supported by the National Research Foundation, Singapore under its AI Singapore Programme (AISG Award No: AISG-RP-2018-003), and the MoE AcRF Tier 2 grant (MOE-T2EP20220-0007) and MoE AcRF Tier 1 grants (RG14/22, RG95/20).

{\small
\bibliographystyle{ieee_fullname}
\bibliography{egbib}
}

\clearpage
\begin{appendices}

\twocolumn[{%
\section*{\LARGE Appendix}

\renewcommand\twocolumn[1][]{#1}%

\begin{center}
\vspace{-0.1in}
    \centering
    \captionsetup{type=figure}
    \includegraphics[width=0.95\textwidth]{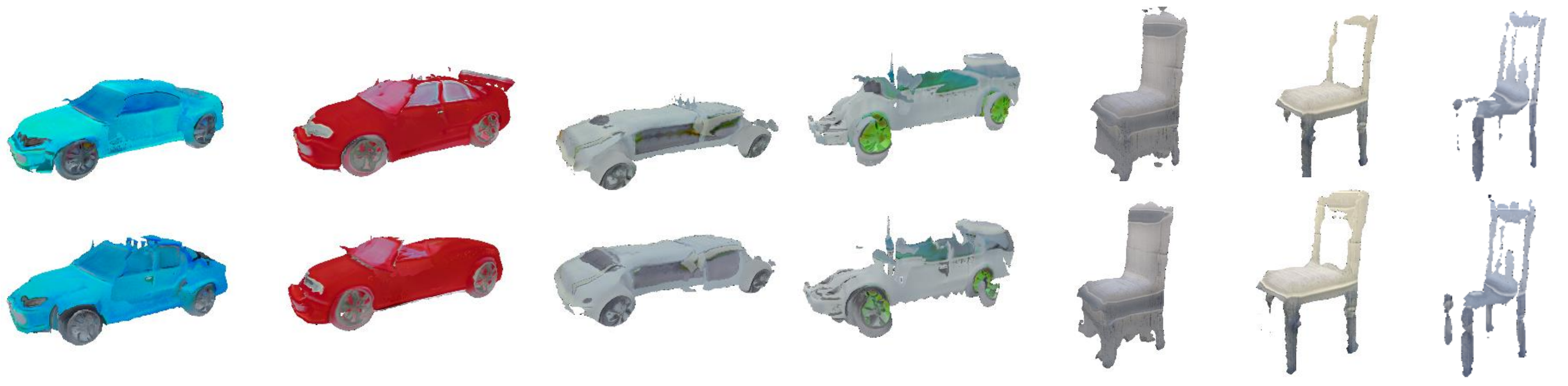}
    \captionof{figure}{We show the generated results of the model trained from scratch, where we update all model parameters including the mapping networks, generator, and discriminators. We observe that the model easily collapses during the training.}
    \label{fig:all_vs_part}
\end{center}%
\begin{center}
\vspace{-0.1in}
    \centering
    \captionsetup{type=figure}
    \includegraphics[width=0.95\textwidth]{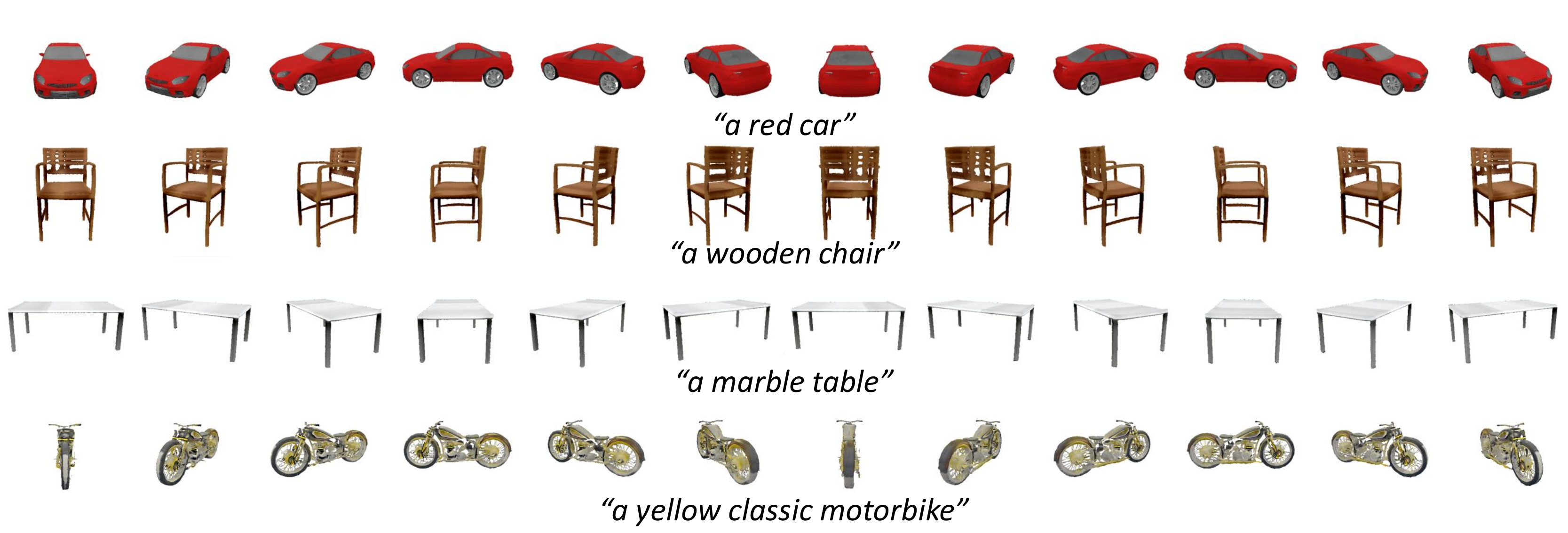}
    \captionof{figure}{We show the multi-view results of generated objects via neural rendering. Our method produces multi-view consistent results related to the input texts.}
    \label{fig:multiview}
\end{center}%
}]

\section{Comparison of different training strategies}
In \cref{fig:all_vs_part}, we show the results of training the entire model from scratch with both the CLIP loss and the GAN loss. 
We observe that the model tends to collapse when we update the entire network. More loss functions involved during the training phase make it harder to converge for the generator and discriminators, resulting in coarse and incomplete generation outputs. While updating the mapping network only retains the pretrained network capability, which is learned during the unconditional training. 

\begin{figure*}[]
\begin{center}
\includegraphics[width=0.9\textwidth]{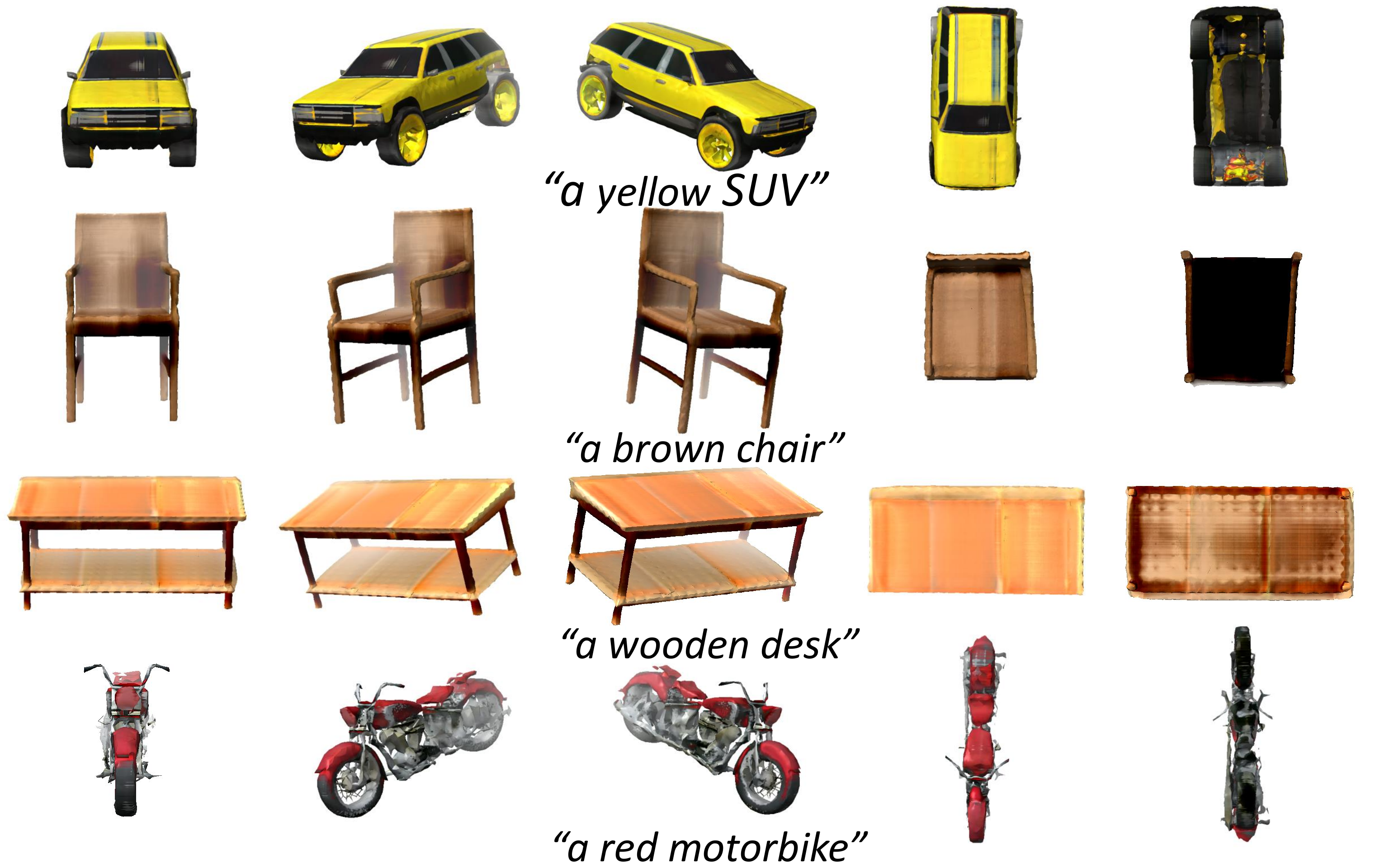}
\end{center}
  \caption{We show the multi-view visualizations of the generated textured meshes, given the input texts. We also present the top views and bottom views of the generated meshes. These mesh results are visualized with ChimeraX\cite{goddard2018ucsf}.}
\label{fig:multiview_mesh}
\end{figure*}

\begin{figure*}[]
\begin{center}
\includegraphics[width=0.95\textwidth]{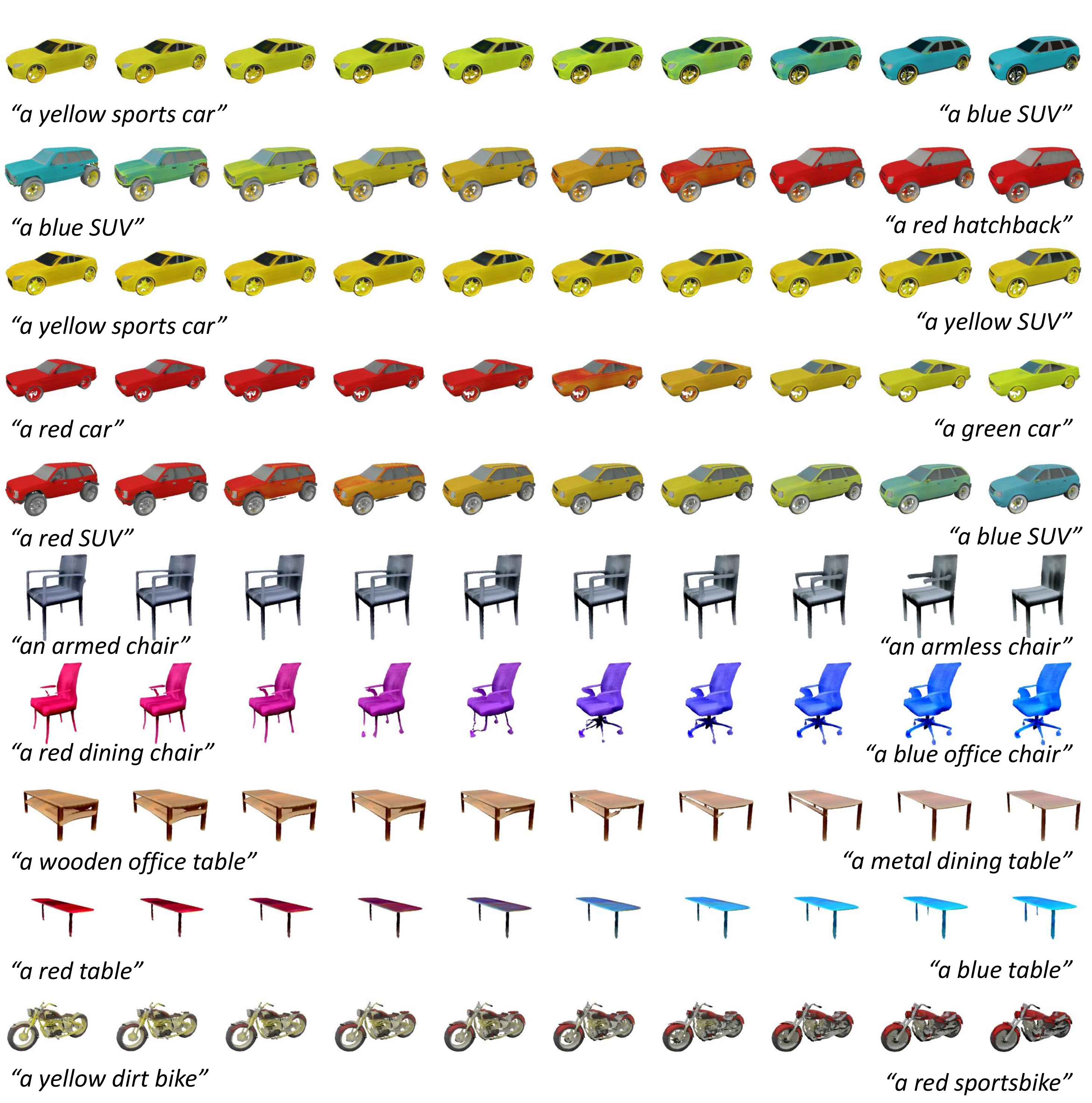}
\end{center}
\vspace{-10pt}
  \caption{We show the interpolation results between two text inputs. Note that each row shares the same sampled random noise. For each $\langle$source, target$\rangle$ pair, we use the same sampled random noise vector $\z$, and their corresponding CLIP text features to generate the latent codes $\w$. Then, the interpolation is performed between the source and target latent codes $\langle \w_1, \w_2 \rangle$. The interpolated latent codes are fed into the generator to synthesize the results.}
\label{fig:interpolation}
\vspace{-10pt}
\end{figure*}

\section{Details of the mapping networks}
We follow \cite{karras2019style,gao2022get3d} to implement the text-conditioned mapping networks, in which we take the random vectors $\z \in \mathbb{R}^{512}$ and CLIP text features $E_t(t) \in \mathbb{R}^{512}$ as input.
We first adopt 2-layer MLPs $f_{t}$ to map $E_t(t)$ to another space so that we can concatenate $E_t(t)$ with the random vectors $\z$. Then we produce latent codes $\w \in \mathbb{R}^{512}$ from the concatenation of $\langle f_t(E_t(t))$, $\z \rangle$ by 8-layers MLPs $f_{map}$, which can be denoted as $\w = f_{map}(\z, f_t(E_t(t)))$.
Please note each layer of the MLPs is a fully-connected layer having 512 hidden dimensions and a leaky-ReLU activation. We use the same architecture for the geometry and texture mapping networks.

\section{More qualitative results}
\textbf{Multi-view consistency.}
In \cref{fig:multiview}, we show that our proposed method generates multi-view consistent images from neural rendering with respect to the input text prompts. In \cref{fig:multiview_mesh}, we visualize the generated textured meshes in different views. The multi-view visualizations illustrate that our model can generate 3D shapes that are consistent with the input texts on each side of the objects.

\textbf{Interpolation Results.}
We produce interpolation results between two input texts in \cref{fig:interpolation}. In each row, we use the same sampled random noise $\z$ with different input texts to generate the source and target latent codes $\w_1$ and $\w_2$. Then, we produce the interpolation results between $\w_1$ and $\w_2$. We show that our learned mapping networks generate smooth and meaningful latent codes from the text input to guide shape generation.

\end{appendices}

\end{document}